\documentclass[10pt,twocolumn,letterpaper]{article}

\usepackage[,algorithms]{wacv}      

\usepackage{graphicx}
\usepackage{amsmath}
\usepackage{amssymb}
\usepackage{booktabs}
\usepackage{times}
\usepackage{epsfig}

\usepackage[accsupp]{axessibility}

\usepackage{booktabs}

\usepackage{ifsym}
\usepackage{listings}
\usepackage[normalem]{ulem}
\usepackage{marvosym}
\usepackage{soul}

\usepackage{threeparttable}
\usepackage{multirow}
\usepackage{pifont}
\newcommand{\cmark}{\ding{51}}%
\newcommand{\xmark}{\ding{55}}%
\usepackage[pagebackref,breaklinks,colorlinks]{hyperref}

\usepackage[capitalize]{cleveref}
\crefname{section}{Sec.}{Secs.}
\Crefname{section}{Section}{Sections}
\Crefname{table}{Table}{Tables}
\crefname{table}{Tab.}{Tabs.}

\def\confName{WACV}
\def\confYear{2024}

\begin{document}

\title{Annotation-free Audio-Visual Segmentation}


\author{Jinxiang Liu$^{1}$, \ Yu  Wang$^{1,2}$, \ Chen Ju$^1$, \  Chaofan Ma$^1$,  Ya Zhang$^{1,2} \textsuperscript{\Letter}$, \ Weidi Xie$^{1,2}\textsuperscript{\Letter}$\\[3pt]
$^1$ Cooperative Medianet Innovation Center, Shanghai Jiao Tong University \ \
$^2$ Shanghai AI Laboratory\\
{\tt\small \{jinxliu,\,yuwangsjtu,\,ju\_chen,\,chaofanma, ya\_zhang,\,weidi\}@sjtu.edu.cn}
}

\twocolumn[{%
\renewcommand\twocolumn[1][]{#1}%
\maketitle
\vspace{-0.5cm}
\begin{center}
    \centering
    \includegraphics[width=1.0\linewidth, trim=0 18 0 0]{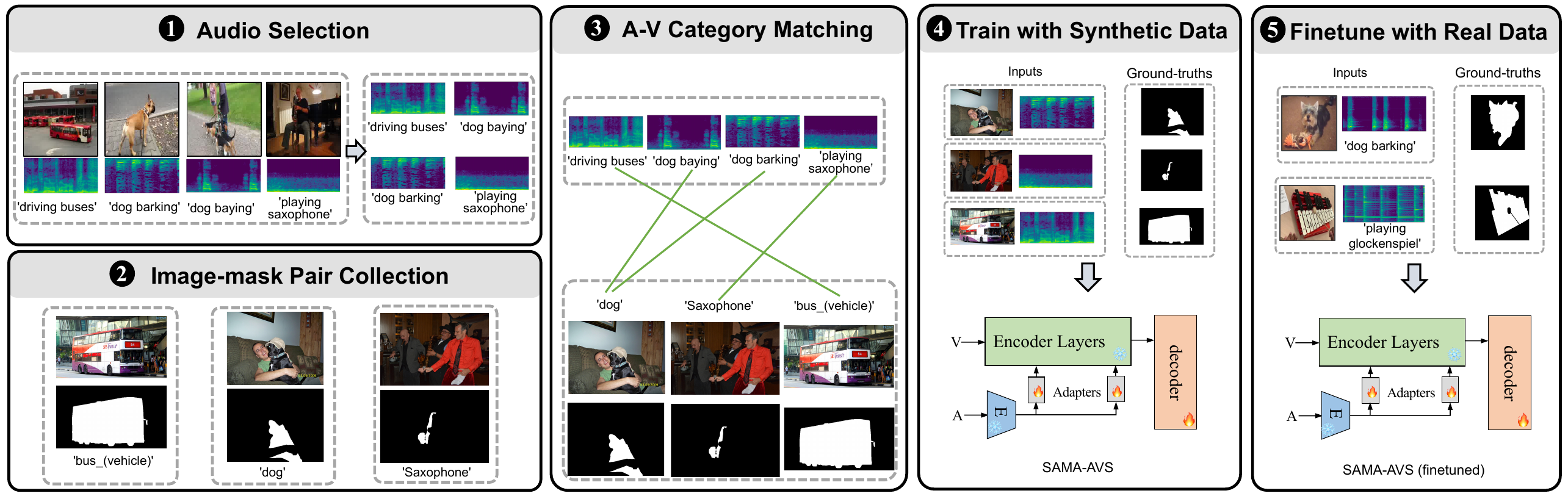}
    \captionof{figure}{
    We propose an ingenious pipeline to synthesize datasets for audio-visual segmentation (AVS), by leveraging \textit{off-the-shelf} large image segmentation datasets and audio corpus. 
    The pipeline is \textit{annotation-free} and can easily be \textit{scaled up} to cover more categories.
    Besides, we develop an effective method SAMA-AVS to adapt \textit{segmentation anything}   (\textbf{SAM}) with \textbf{A}dapters to the \textbf{AVS} task. By introducing only a small number of model parameters, the model performance is substantially enhanced and achieves state-of-the-art performance.
    }
\label{fig:syn-pipeline}
\end{center}%
}]
 
\begin{abstract}
The objective of Audio-Visual Segmentation (AVS) is to localise the sounding objects within visual scenes by accurately predicting pixel-wise segmentation masks.
To tackle the task, it involves a comprehensive consideration of both the data and model aspects. In this paper, first, we initiate a novel pipeline for generating artificial data for the AVS task without extra manual annotations. 
We leverage existing image segmentation and audio datasets and match the image-mask pairs with its corresponding audio samples using category labels in segmentation datasets, that allows us to effortlessly compose (image, audio, mask) triplets for training AVS models. 
The pipeline is annotation-free and scalable to cover a large number of categories. 
Additionally, we introduce a lightweight model SAMA-AVS which adapts the pre-trained segment anything model~(SAM) to the AVS task. 
By introducing only a small number of trainable parameters with adapters, the proposed model can effectively achieve adequate audio-visual fusion and interaction in the encoding stage with vast majority of parameters fixed. 
We conduct extensive experiments, and the results show our proposed model remarkably surpasses other competing methods. Moreover, by using the proposed model pretrained with our synthetic data, the performance on real AVSBench data is further improved, achieving 83.17 mIoU on S4 subset and 66.95 mIoU on MS3 set.
The project page is \href{https://jinxiang-liu.github.io/anno-free-AVS/}{https://jinxiang-liu.github.io/anno-free-AVS/}.
\end{abstract}


\section{Introduction}\label{sec:intro}

Audio-visual segmentation (AVS) task involves segmenting the sounding objects corresponding to the audio cues in the video frames. 
Previous research has explored this task under the umbrella of self-supervised learning by leveraging the occurrence of audio-visual signals~\cite{hu2021class,hu2020discriminative,hu2019deep,lin2021unsupervised,chen2021localizing,song2022self,liu2022exploiting,senocak2018learning,rouditchenko2019self}. 
However, these training objectives lack fine-grained pixel-wise supervision and results in coarse segmentations. 
This poses limitation to the applicability of AVS in real-world scenarios where accurate segmentations are required, such as video surveillance, multi-modal video editing and the robotics industry.
Recently, Zhou~\etal~\cite{zhou2022avs} tackled the AVS problem through supervised training, and they manually annotated a video dataset with pixel-level segmentations for sounding objects in the video frames. 
However, due to the expensive annotation costs, the dataset has a deficiency in the limited sample quantity and diversity.

To mitigate this, we propose an {\em annotation-free} and {\em scalable} pipeline to construct artificial datasets for the AVS task.
Strictly speaking, an ideal model for the task requires \textit{dual-level} recognition ability: semantic level and instance level.
For example, given the video frames with the sound of dog barking, the model should first distinguish the dogs from the cats with the dog barking sound (\textit{semantic level}); and then further resolve the instance ambiguity by identifying the specific dog that barks when there are multiple dogs present (\textit{instance level}).
Yet in practice, the ability to distinguish semantic categories may have been sufficient for localising sounding objects in most daily scenarios, which can be achieved by training on artificially constructed data with semantic-identical image and audio pairs.
Based on such observation, we present a practical pipeline to artificially synthesize AVS datasets by leveraging off-the-shelf image segmentation and audio datasets without additional manual annotations, resolving the issue of insufficient training data for the AVS task. 

In Figure~\ref{fig:syn-pipeline}, we depict the proposed pipeline to construct a synthetic dataset by leveraging off-the-shelf image segmentation and audio datasets. For instance, we can combine an image-mask pair labeled as ``dog'' from the image segmentation dataset LVIS with an audio sample labeled as ``dog barking'' from the audio dataset VGGSound to obtain an (\textit{image, mask, audio}) triplet, which serves as a training sample for the AVS task. 
In this triplet, \textit{image} and \textit{audio} are the model's inputs, and \textit{mask} provides supervision for model training. 
Comparing to the existing human-annotated dataset AVSBench~\cite{zhou2022avs}, this dataset synthesis pipeline inherits existing free annotations from computer vision communities, such as LVIS~\cite{gupta2019lvis}, Open Images~\cite{kuznetsova2020open}; and most importantly, it requires zero extra annotation, 
thus can easily be scaled up to cover a large number of categories.

In addition, the foundation model \textit{segment anything}~\cite{kirillov2023segment} (SAM) for image segmentation has attracted great attention in computer vision recently.
SAM is pretrained with large-scale data and has demonstrated strong generalisation ability. 
In this paper, we first evaluate two variants of SAM on the AVS task.
One is directly applying SAM to the task, the other is the proposed Audio-Prompted SAM (AP-SAM) which encodes the audios as the prompts for the mapping inputs of the SAM decoder.
However, both models performs suboptimally.
We analyze that vanilla SAM does not associate the semantics between audios and images in the pretraining stage; and even incorporating the audio prompts in the later stage of decoder part (AP-SAM), the fusion of audio-visual features is too \textit{shallow} to facilitate the AVS task due to the light-weight design of the SAM decoder.
Given such finding, a natural question arise: \textit{how can we exploit the powerful ability of SAM, and effectively adapt it to the AVS task?}
In this paper, we propose SAMA-AVS to employ the adapters to transfer the powerful knowledge from SAM for AVS task.
In a nutshell, we utilise the adapters to perform audio-visual fusion in the early stage and only train the adapters with much fewer parameters in the encoding stage.
Thereby, the knowledge of SAM is employed and the audios features can be injected and fused with visual features efficiently and effectively in the early stage while only introducing a small number of parameters (\textless 0.4\% of the entire model). And experimental results show SAMA-AVS outperforms other SAM-based methods by a large margin and achieves state-of-the-art performance on the AVSBench dataset, indicating the effectiveness of the proposed method.

To summarize, the contributions are as follows:

\vspace{-0.15cm}
\begin{itemize}
\setlength\itemsep{0.1cm}
    \item We propose a novel annotation-free pipeline to synthesize datasets for the AVS task, by harnessing the existing image segmentation datasets and audio corpus. The pipeline circumvents the cost of manual annotation and thus easily scales up to cover a greater number of categories for the task.
    \item As an instantiation of the pipeline, we present the AVS-Synthetic dataset, which is significantly larger and more diverse than existing AVS datasets. We conduct experiments with the synthetic dataset and demonstrate its strong zero-shot and few-shot transfer ability to real-world scenarios.
    \item We devise a model termed SAMA-AVS for AVS task based on SAM foundation model. To realise the deep fusion of audio-visual representations, the method deploys simple adapters to inject the audio information into the pretrained SAM, and allows the deep audio-visual fusion in the encoding stage with the vast majority of parameters fixed. Extensive experiments demonstrate the effectiveness of our proposed method. 
\end{itemize}

\section{Related Work}
\label{sec:related}
\noindent \textbf{Audio-visual segmentation. }
In the recent literature, researchers have explored a range of audio-visual tasks with the goal of getting a comprehensive understanding of multimedia resources. These tasks include audio-visual sound separation~\cite{zhao2018sound,zhao2019sound,gao2021visualvoice,ephrat2018looking,tzinis2022audioscopev2}, 
visual sound source localisation~\cite{hu2021class,hu2020discriminative,hu2019deep,lin2021unsupervised,chen2021localizing,song2022self,liu2022exploiting,senocak2018learning,qian2020multiple} and audio-visual video understanding~\cite{kazakos2019epic,tian2020unified,lin2019dual,lee2020cross,tian2020unified}.
Visual sound source localisation (VSSL), or audio-visual segmentation (AVS), 
aims to localise and segment the sounding object regions based on its audio signals.
Existing approaches~\cite{owens2018audio,zhou2016learning,senocak2018learning,rouditchenko2019self,chen2021localizing,liu2022exploiting} to this problem are primarily based on the co-occurrence of audio and visual signals, which provides only weak instance-level supervision, making it difficult to predict fine-grained pixel-level segmentation masks.
To overcome this issue, Zhou~\etal\cite{zhou2022avs} presented a dataset with mask annotations for sounding objects in the video frames, which is the first benchmark for the supervised AVS task.

\vspace{3pt}
\noindent \textbf{Segment Anything.} SAM~\cite{kirillov2023segment} is released as the first foundation model for image segmentation in computer vision community.
The model is pretrained with over 11 million images associating with over one billion segmentation masks.
It also has the flexibility of supporting diverse prompts including points, boxes, masks and texts.
With the advantages of SAM, it is soon utilised to attack various visual problems including medical image segmentation~\cite{DBLP:journals/corr/abs-2304-12620,DBLP:journals/corr/abs-2305-00035,DBLP:journals/corr/abs-2306-06370,DBLP:journals/corr/abs-2304-14674,DBLP:journals/corr/abs-2306-00499,DBLP:journals/corr/abs-2304-12306}, weakly-supervised semantic segmentation~\cite{DBLP:journals/corr/abs-2305-11003,DBLP:journals/corr/abs-2305-01275,DBLP:journals/corr/abs-2305-05803}, few-shot segmentation~\cite{DBLP:journals/corr/abs-2305-03048,DBLP:journals/corr/abs-2305-13310}, 3D vision~\cite{DBLP:journals/corr/abs-2304-12308,DBLP:journals/corr/abs-2306-09347}, shadow detection~\cite{chen2023sam,DBLP:journals/corr/abs-2306-06113,DBLP:journals/corr/abs-2305-16698}, camouflaged object segmentation~\cite{chen2023sam,DBLP:journals/corr/abs-2304-05750,ji2023sam} and so on.
Currently, few works except~\cite{mo2023av} explores SAM for the AVS task but the results from ~\cite{mo2023av} are less satisfying.
In this paper, we examine the performance of vanilla SAM and describe our proposed SAM-based models for the AVS task.

\vspace{3pt}
\noindent \textbf{Adapter.} Adapter~\cite{houlsby2019parameter} has become one of  the most prevalent methods in parameter-efficient transfer learning.
By introducing only a small number of parameters, it can effectively leverage the knowledge within the pretrained models and transfer to the related problems, thus it is now widely utilised for both NLP tasks~\cite{houlsby2019parameter,karimi2021compacter,sung2022lst,alayrac2022flamingo,liu2022few,stickland2019bert} and CV tasks~\cite{li2022exploring,chen2023sam,lin2022frozen,rebuffi2018efficient,chen2022vision,liu2023explicit}. 
Recently, it is also employed to solve multi-modal problems~\cite{lee2020parameter,sung2022vl,lin2023vision}.
\section{Dataset}\label{sec:method}
Here, we start by detailing the dataset construction procedure in Sec.~\ref{sec:pipeline}; then we present the statistics of the proposed AVS-Synthetic dataset in Sec.~\ref{sec:avs-synthetic}.

\subsection{Dataset Synthesis Pipeline}
\label{sec:pipeline}

Existing datasets for audio-visual segmentation,
{\em e.g.}, AVSBench~\cite{zhou2022avs},
are all collected by manually annotating frames with pixel-wise segmentation masks, which is often expensive and only covers a limited number of categories (only 23).
To remedy this, we initiate a scalable pipeline for constructing a synthetic dataset, termed as \textbf{AVS-Synthetic} dataset.
The dataset construction procedure leverages existing image segmentation  datasets such as LVIS~\cite{gupta2019lvis}, Open Images~\cite{kuznetsova2020open}, and audio classification dataset, such as VGGSound~\cite{chen2020vggsound}. Therefore it is free of extra manual annotations and can easily be scaled up. 

Figure~\ref{fig:syn-pipeline} illustrates our proposed pipeline for building a synthetic dataset for semantic-level audio-visual segmentation. Generally speaking, it consists of three steps:

\vspace{3pt}
\noindent \textbf{(1) Audio Selection.}
We select audios using the instance-level class labels from publicly-released audio dataset. In this paper, we employ VGGSound~\cite{chen2020vggsound}, which is a large-scale audio-visual dataset collected using computer vision techniques, as a data source for providing audio sample clips in our proposed AVS-Synthetic dataset. However, other audio datasets such as Audio Set~\cite{gemmeke2017audio} can also be used. 

\vspace{3pt}
\noindent \textbf{(2) Image-mask Pair Collection.}
Next, based on the categories of selected audio clips, we pick the categorized image and segmentation mask pairs for the objects in the target categories. This can be easily achieved by filtering the existing popular image segmentation datasets such as LVIS~\cite{gupta2019lvis} and Open Images~\cite{kuznetsova2020open}.

\vspace{3pt}
\noindent \textbf{(3) Audio-Visual Category Matching.}
Till here, we have separately collected the audio and visual samples.
We match each image-mask pair with their corresponding audio clip, thus forming the (\textit{image, mask, audio}) triplet as a training sample for AVS task.
For each triplet, (\textit{image, audio}) serve as model inputs and the \textit{mask} is the ground-truth segmentation mask to supervise the model training.
Matching these pairs is done by using the class labels of both image-mask pairs and the audio labels. 
Notably, different datasets may have different class labels while referring to the same thing; such as \textit{computer keyboard}: ``computer\_keyboard'' (LVIS) and ``typing on computer keyboard'' (VGGSound); \textit{dog}: ``dog'' (LVIS), ``dog barking'' (VGGSound) and ``dog baying'' (VGGSound). Minimal manual verification is involved in this matching procedure.

\subsection{AVS-Synthetic Statistics}\label{sec:avs-synthetic}
As an instantiation of our \textit{annotation-free} dataset collection pipeline for AVS, \textbf{AVS-Synthetic} dataset is proposed that covers 62,609 sounding object instances spanning 46 common categories.
The training set size is 52,609, and the sizes of validation set and test set are both 5,000. A detailed comparison between AVS-Synthetic and the human annotated AVSBench datasets~\cite{zhou2022avs} is listed in Table~\ref{tab:datasets}.

\begin{table}[!htb]
\footnotesize
\centering
\setlength{\tabcolsep}{2.5pt}
\label{tab:avs-syn-stat}
\begin{tabular}{lcrrcc}
\toprule
Dataset & Classes & Images & Masks & Data.Source & Human   \\
\midrule
AVSBench~\cite{zhou2022avs} & 23 & 4932 & 10,852   & \cite{chen2020vggsound}   &  \cmark\\
\textbf{AVS-Synthetic~(ours)} & 46 & 58,405 & 62,609 & \cite{gupta2019lvis,kuznetsova2020open,chen2020vggsound}   & \xmark    \\
\bottomrule
\end{tabular}
\vspace{-4pt}
\caption{Comparison between our proposed AVS-Synthetic and AVSBench (S4)~\cite{zhou2022avs} datasets. 
AVSBench is collected with human annotations. }
\label{tab:datasets}
\vspace{-0.2cm}
\end{table}
\section{Method}
In Sec.~\ref{sec:problem}, we start by introducing the problem formulation. Then we describe the two SAM-based models for AVS task in Sec.~\ref{sec:revisit-sam} and Sec.~\ref{sec:apsam}.
Next, we detail the proposed SAMA-AVS architecture in Sec.~\ref{sec:sama-avs}. Finally, we describe the model training in Sec.~\ref{sec:objectives}.

\subsection{Problem Formulation}
\label{sec:problem}
For the task of audio-visual segmentation (AVS), 
the input data consists of a sequence of video frames, $\mathcal{V}=\left\{v_i\right\}_{i=1}^T$, where $v_{i} \in \mathbb{R}^{3 \times H_{0} \times W_{0}}$, and its corresponding audios $\mathcal{A}=\left\{a_i\right\}_{i=1}^T$, 
where $a_{i} \in \mathbb{R}^{H_a\times W_a}$ refers to the corresponding audio spectrogram. The objective of AVS is to segment the sounding objects in each frame based on the acoustic cues. The target segmentation can be binary masks, {\em i.e.}, $\mathcal{M}=\left\{M_i\right\}_{i=1}^T$, and $M_{i} \in  \{0,1\}^{{H_{0} \times W_{0}}}$.

\subsection{Revisiting SAM as Baseline}\label{sec:revisit-sam}
Segment Anything (SAM)~\cite{kirillov2023segment} is proposed as a foundation model for scene segmentation, which is pretrained on 11M images with 1 billion segmentation masks. In terms of the architecture design, SAM consists of three components: an image encoder, a prompt encoder and a light-weight mask decoder.
The image encoder employ the Vision Transformer~\cite{sharir2021image} pretrained with MAE~\cite{he2022masked} to extract the image features with the shape of $C\times H\times W$ from the input images.
For the prompt encoder, its supported prompt inputs can be divided into two categories: sparse prompts including points, boxes, texts, and masks.
The points and boxes are processed by a combination of positional encodings and learned embeddings. The texts are the embeddings from the text encoder of CLIP~\cite{radford2021learning}. The mask are summed into the image embeddings with convolution operations.
Note that, when no prompts are given, SAM can also generate the segmentation proposals by traversing the entire image. 

In this paper, we first investigate the performance of the vanilla SAM model on the AVS task,
by picking the predicted mask that best matches the ground-truth~\cite{tang2023can,ji2023sam}. Concretely, for an image $I$ with the ground-truth mask $G$, SAM model predicts $N$ masks $\{M_i\}_{i=1}^N$. 
Then by calculating the intersection over union (IoU) of the ground-truth mask with the predictions, we select the predicted mask with the highest IoU value as the final prediction of SAM for evaluation. This method serves as the baseline for our proposed SAM-based methods. 

\subsection{Audio-Prompted SAM}\label{sec:apsam}
Inspired by the way of dealing with text prompts of vanilla SAM as elaborated in Sec.~\ref{sec:revisit-sam}, we treat the audio as another type of prompt by extracting the audio embeddings and set up another SAM-based model named Audio-Prompted SAM (AP-SAM) for the AVS task.
Concretely, given an audio clip, we first compute its features with the VGGish~\cite{hershey2017cnn} model pretrained on AudioSet~\cite{gemmeke2017audio} dataset, then spatially pool the features into a vector embedding.
Similar to the SAM design, we pose the audio features as the prompt tokens.
And thus the image embedding ($256\times64\times64$) from the image encoder and the audio prompt tokens ($1\times256$) are passed to the light-weight mask decoder.
The mask decoder updates the image embedding and audio prompt tokens, and uses the output tokens to perform mask prediction, following the same design of vanilla SAM. 
At training phase, we fix the parameters of both image encoder and audio encoder to preserve their knowledge, and only update the weights within the mask decoder.

\subsection{SAMA-AVS: SAM with Adapters for AVS}\label{sec:sama-avs}
In the AP-SAM model, although the image embedding and the audio embedding tokes are interacting with each other in the mask decoder, we conjecture the fusion is not adequate to drive the AVS task.
For the mask decoder of SAM, it is light-weighted with only two transformer decoder layers.
The reason why it works with such simple design is that the prompts including points, boxes and masks generally only provide explicit location information of target segmentation.
And the whole models are trained with millions of images with semi-supervised learning.
Therefore, a two-layer mask decoder design is competent enough for the vanilla SAM pretraining. 
However, to complete the audio-visual segmentation (AVS) task successfully, it requires the model to understand and associate the semantics between both image and audio embeddings; but this audio-image semantic understanding are not explored in the vanilla pretrained SAM. And the fusion and interaction of audio-visual tokens using the two-layer mask decoder of SAM may be too limited.

\begin{figure}[t]
\centering
\vspace{0.1cm}
\includegraphics[width=.99\linewidth]{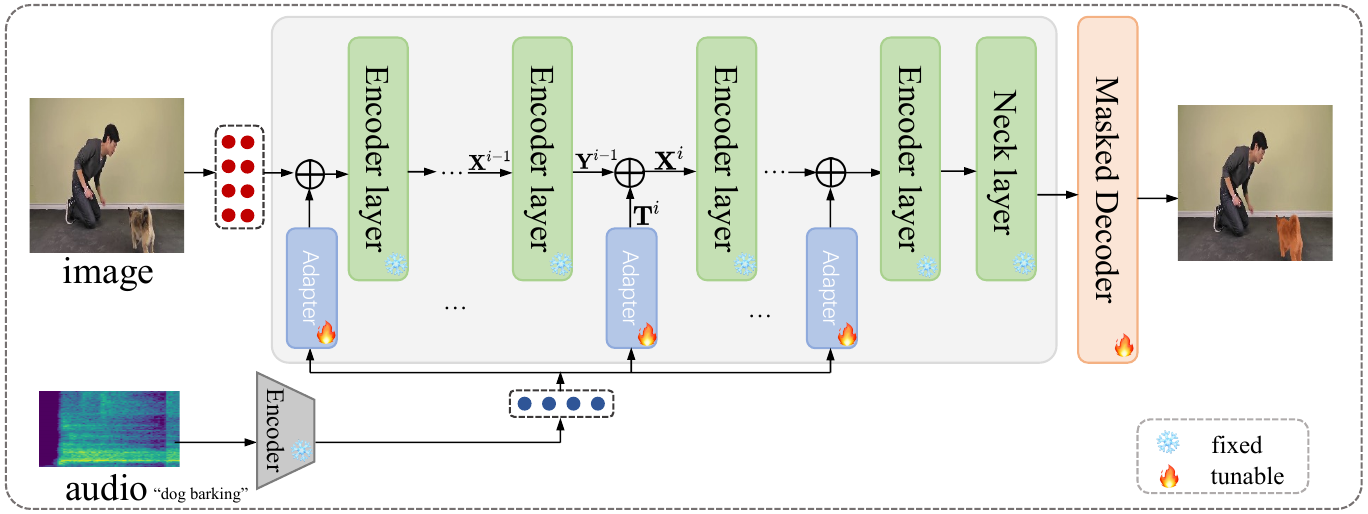}
\vspace{-0.1cm}
\caption{The architecture of the proposed SAMA-AVS.}
\label{fig:archi}
\vspace{-0.4cm}
\end{figure}

One simple solution is to inject the audio information into the image features at the encoding process.
However, the computation cost of finetuning the whole encoder weights would be prohibitively high. 
To tackle the problem, we propose to perform deep audio-visual fusion with the assistance of adapters. 
We term the proposed framework based on SAM with adapters for AVS task as SAMA-AVS. 
We elaborate the method shown in Figure~\ref{fig:archi} in more detail.

\vspace{3pt}
\noindent \textbf{Image tokenization.}
Our model accepts the visual frames and audio as the inputs.
For visual input, given a video frame $I\in \mathbb{R}^{3\times H\times W}$, we first upsample the image to the shape of ${3\times 1024\times 1024}$ to meet the requirements of SAM.
Then we patchify the image into non-overlapping patches and flatten them into visual embeddings.
Before feeding the features into the network, we add the visual embeddings with learnable position embeddings initialized with zeros. 
Finally we obtain the visual embeddings $X_v \in \mathbb{R}^{N\times D}$.

\vspace{3pt}

\noindent \textbf{Audio tokenization.}
For the audio, we obtain the audio spectrogram $A\in \mathbb{R}^{H_a\times W_a}$ of one second audio clip corresponding to the video frame.
we compute its features with the pretrained VGGish~\cite{hershey2017cnn} model, then spatially pool the features into a vector embedding $X_a\in \mathbb{R}^\ell$ .

\vspace{3pt}
\noindent \textbf{Deploying adapters at encoding stage.}
The adapters are responsible for injecting the audio information into the feature encoding process.
Specifically, the audio tokens are transformed with learnable parameters and then added to the outputs of each transformer encoder layer. 
Formally, we denote the outputs of the $(i-1)$th layer of transformer encoder as $\mathbf{Y}^{i-1}$ and the outputs of $i$th adapter as $\mathbf{T}^i$, then the inputs for the $i$th encoder layer $\mathbf{X}^i$ is obtained by:
\begin{equation}
    \mathbf{X}^{i} = \mathbf{Y}^{i-1} + \mathbf{T}^i.
\end{equation}
After N layers of transformer encoder with adapters, the obtained deeply-fused audio-visual embeddings are fed to the last neck layer to adjust the channel dimension.
Finally the embeddings from the neck layer are fed to the mask decoder to perform mask prediction.

In terms of adapter design, it is simply implemented with two-layer MLPs to adjust the feature dimension and perform feature extraction. 
During the training phase, we fix the parameters of the transform encoder layers and the audio encoder, and only update the parameters of the adapters and the light-weight mask decoder. 
Note that, the parameters introduced by the adapters constitute only {\bf 0.4\%} of the entire model, while yielding significant performance gains, as indicated in the experiments.

\subsection{Training Objectives}\label{sec:objectives}
We use a combination of two commonly used objectives at training time: the BCE (Binary Cross-Entropy) loss and the IoU (Intersection over Union) loss. 
The total loss for optimizing the model is the weighted sum of the BCE loss and IoU loss:
\begin{equation}
    \mathcal{L}_{\text{total}} = \mathcal{L}_{\text{BCE}} + \lambda_{\text{IoU}} \mathcal{L}_{\text{IoU}},
\end{equation}
where $\lambda_{\text{IoU}}$ is weight to balance the losses.

\section{Experiments}

\subsection{Datasets} 
AVSBench~\cite{zhou2022avs} is a newly released dataset for audio-visual segmentation that comes with human-annotated mask annotations. This dataset is divided into two subsets: the semi-supervised Single Sound Source Segmentation (S4) and the fully supervised Multiple Sound Source Segmentation (MS3).

\vspace{4pt}
\noindent \textbf{AVSBench-S4 dataset.}
This subset contains videos with a maximum of one sounding object. 
During training, only the first frame of each video sequence is annotated with a ground-truth segmentation binary mask. At inference time, the goal is to segment the sounding objects across all frames in the video. The subset includes 3,452$/$740$/$740 videos in the training, validation, and test sets, respectively, with a total of 10,852 annotated frames.

\vspace{4pt}
\noindent \textbf{AVSBench-MS3 dataset.}
The MS3 subset consists of videos that may contain multiple sounding objects. All video frames are annotated with masks during both the training and evaluation stages. The subset contains 296$/$64$/$64 videos for training, validation and testing separately, with a total of 2,120 annotated frames.

\vspace{4pt}
\noindent \textbf{AVS-Synthetic dataset.} 
As detailed in Sec~\ref{sec:avs-synthetic}, AVS-Synthetic is a dataset collected using our proposed pipeline. The dataset includes 46 categories with over 60K (\textit{image, audio, mask}) sample triplets. To facilitate training and evaluation, the dataset is divided into training, validation and testing splits, which includes 52,609/5,000/5,000 samples separately.

\subsection{Evaluation Metrics}
We quantify the segmentation quality by adopting standard segmentation metrics outlined in~\cite{zhou2022avs}.
Specifically, we use $\mathcal{M}_{\mathcal{J}}$ to denote the mean Intersection-over-Union (mIoU) of ground-truth binary masks and predicted masks. 
We use $\mathcal{M}_{\mathcal{F}}$, known as the F-score, which is the harmonic mean of precision and recall. In both cases, higher values indicate better segmentation performance.

\subsection{Implementation Details}
We choose the released ViT-H SAM model with pre-trained ViT-H/16 image encoder with 32 Transformer encoder layers as weight initialization.
The resolution of the input images are $1024\times 1024$. 
The weight to balance the two losses is set to 1.
To optimize the model parameters, we employ the AdamW~\cite{loshchilov2017decoupled} optimizer with an initial learning rate of $2\times 10^{-4}$ with cosine decay. We implement the models with PyTorch. 
We train the models for 80 epochs with four NVIDIA RTX A6000 GPUs. 
The batch size is set to 8.

\subsection{Effectiveness of Data Synthesis Pipeline}
We train our proposed SAMA-AVS model on the training split of the AVS-Synthetic dataset and present results on three experimental settings: 
standard evaluation on the test split of AVS-Synthetic, 
zero-shot evaluation on the test split of AVSBench, 
and few-shot evaluation on the test split of AVSBench after real data finetuning.

\begin{figure*}[!tb]
\centering
\vspace{0.1cm}
\includegraphics[width=1.0\linewidth]{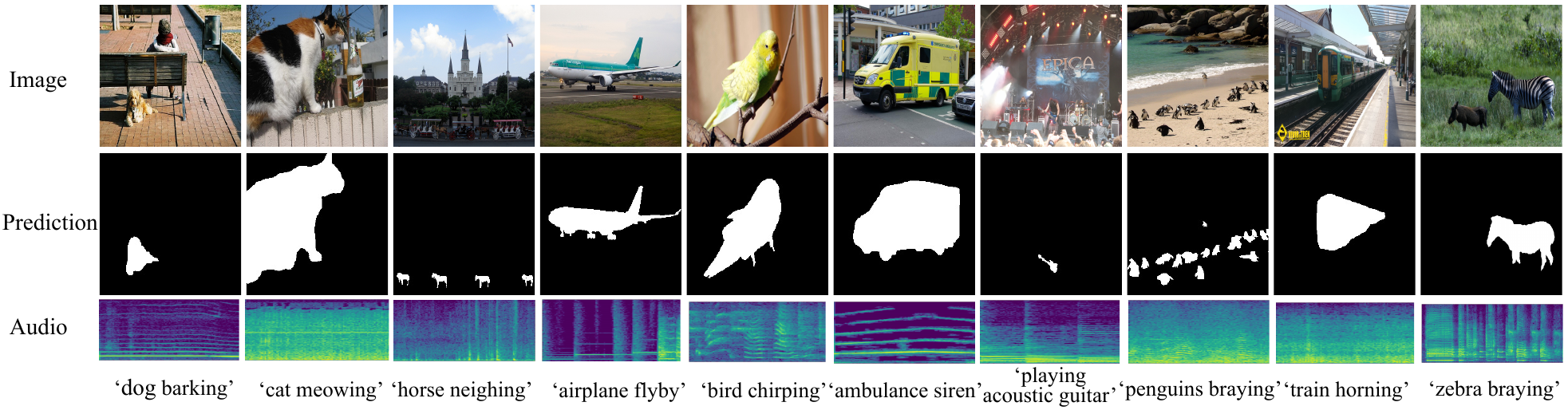}
\vspace{-0.5cm}
\caption{Qualitative results on the test split of AVS-Synthetic. }
\label{fig:syn-val-vis}
\end{figure*}

\begin{table*}[]
\small
\setlength{\tabcolsep}{7.8pt}

\begin{tabular}{ccccccccccc}
\toprule
Trained with & ambulance & cat  & dog  & bus & horse & lion & bird & guitar  & piano  & computer\_keyboard  \\
\midrule
Real    & 73.63 & 83.87 & 84.88  & 85.09 & 81.21  & 91.28 &  85.64 & 85.78 & 82.91  & 85.99    \\
Synthetic & 55.32 & 78.81  & 78.66 & 82.70 &  73.88 & 88.51 & 81.99  & 75.11 & 32.26 & 74.01   \\
\bottomrule
\end{tabular}
\vspace{-5pt}
\caption{Categorized results of direct zero-shot evaluations of the models trained with synthetic data on real test split of AVSBench (S4)~\cite{zhou2022avs}; The ``Real'' row means the model is trained with the real data of AVSBench training split; and the ``Synthetic'' row means the model is trained with AVS-Synthetic. Metric: $\mathcal{M}_{\mathcal{J}}$ (mIoU).}
\label{tab:cat-res}
\vspace{-0.2cm}
\end{table*}

\vspace{3pt}
\noindent \textbf{Results on synthetic dataset.}
We trained the SAMA-AVS model using the training and validation splits of AVS-Synthetic dataset, and then evaluate on the test split.
It achieves the results of 79.21 on $\mathcal{M}_{\mathcal{J}}$ and 0.855 on $\mathcal{M}_{\mathcal{F}}$. 
Figure~\ref{fig:syn-val-vis} provides some qualitative results; and our model can accurately segment the objects semantically corresponding to the audios.

\begin{table}[]
    \setlength{\tabcolsep}{7pt}
    \centering
    \begin{tabular}{ccccc}
    \toprule
    \multirow{2}{*}{Real Data} &  \multicolumn{2}{c}{\textit{w/o} P.T.} & \multicolumn{2}{c}{\textit{w/} P.T.} \\
      \cmidrule(r){2-3} \cmidrule(r){4-5}   &   $\mathcal{M}_{\mathcal{J}}$ &  $\mathcal{M}_{\mathcal{F}}$ & $\mathcal{M}_{\mathcal{J}}$ & $\mathcal{M}_{\mathcal{F}}$ \\
    \midrule
      0\%   & 17.65  & 0.227  & 64.05 ({\color{green}+46.1})  &  0.744   \\ 
      10\%  & 72.76  & 0.813  &  76.22 ({\color{green}+3.46}) & 0.844    \\ 
      20\%  &  75.93 & 0.840  & 79.05 ({\color{green}+3.21})  &  0.870  \\ 
      30\%  & 78.20  & 0.859  & 80.41 ({\color{green}+2.21}) & 0.876     \\ 
     100\%  &  81.53 & 0.886  & \textbf{83.17} ({\color{green}+1.64}) &  \textbf{0.901}   \\ 
     
    \bottomrule
    \end{tabular}
    \vspace{-3pt}
    \caption{Finetuning the model with different percentages of real training data from AVSBench dataset. ``\textit{w/} P.T.'' refers to the model pretrained on AVS-Synthetic. 
0\%, 10\%, 20\% and 30\% are the ratios of real data from AVSBench S4 training set; while ``\textit{w/o} P.T.'' means the normal training without using pretrained weights as initialization.
}
    \label{tab:real-ft}
\vspace{-0.5cm}
\end{table}

\vspace{3pt}
\noindent \textbf{Zero-shot results on real dataset.}
In this section, we examine the zero-shot transfer setting.
Concretely, we train the model on AVS-Synthetic dataset and test it on the AVSBench S4 test set. 
Table~\ref{tab:cat-res} quantitatively showcases the results for the overlapping categories between AVS-Synthetic and AVSBench. 
Surprisingly, the model achieves remarkably close results on certain categories, 
such as ``cat'', ``bus'', ``bus'' and ``lion'',
demonstrating  the effectiveness of our proposed data synthesis procedure, 
that enables zero-shot transfer to real data.
We also observe the model trained with synthetic data generally lags behind the one trained with real data for all categories, which may be attributed to the domain gap between different image/video datasets. The video of the AVSBench dataset are from VGGSound and tend to be object-centric, while the image sets of AVS-Synthetic comprise LVIS and Open Images which are scene-centric. 
In future work, further domain adaptation techniques should be investigated to alleviate this issue.

\vspace{3pt}
\noindent \textbf{Data-efficient adaptation on real data.}
To narrow the gap between synthetic and real data, 
we adopt a small number of real data from the AVSBench-S4 subset,
and fine-tune the model trained with AVS-Synthetic, 
resembling the data-efficient few-shot learning scenario. 
The results in Table~\ref{tab:real-ft} demonstrate that the model trained on AVS-Synthetic achieves 64.05 mIoU ($\mathcal{M}_{\mathcal{J}}$) without using any real data. However, by using only 10\% of real data for fine-tuning, the performance can be significantly improved to 76.22 mIoU ($\mathcal{M}_{\mathcal{J}}$). 
We observe \textit{consistent} improvement when more real data (20\%/30\%/100\%) is provided for training. 
And the model (w/ P.T.) trained using 10\% real data outperforms the model (w/o P.T.) using 20\% real data.
Similarly, the model (w/ P.T.) trained using 20\% real data outperforms the model (w/o P.T.) using 30\% real data.
The results show that synthetic dataset can effectively reduce the demand for annotating real data with its better performance.
And when using 100\% real data, the model using the weights of AVS-Synthetic pretrained model improves the performance by 1.64 $\mathcal{M}_{\mathcal{J}}$, achieving 83.17 $\mathcal{M}_{\mathcal{J}}$ and 0.901 $\mathcal{M}_{\mathcal{F}}$.

Overall, these results demonstrate that the model trained with synthetic data from our proposed pipeline brings two benefits: (i) improving real-data efficiency, 
(ii) boosting the model's overall performance for the AVS task.

\subsection{Effectiveness of SAMA-AVS}
In this part, we conduct experiments to compare the performance of our proposed SAMA-AVS with other SAM-based methods in Sec.~\ref{sec:com-sambase} and other state-of-the-art methods in Sec.~\ref{sec:com-sotas}. 
Ablation study is done in~\ref{sec:abla}.
We also present some segmentation results qualitatively in Sec.~\ref{sec:quali}.
To ensure fair comparison, 
we do not utilize synthetic data pretraining unless otherwise stated.

\subsubsection{Comparison with SAM-based Baselines.}\label{sec:com-sambase}
We evaluate the performance of different methods on AVSBench, 
including SAM, AV-SAM~\cite{mo2023av}, AP-SAM and SAMA-AVS.
For the SAM baseline, we load the released model weights of ViT-H SAM model without further training and adopt the maximum segmentation evaluation as described in Sec.~\ref{sec:revisit-sam}.
For the AV-SAM~\cite{mo2023av} method, we cite the results from its original paper. 
For the proposed AP-SAM and SAMA-AVS, we use the same experimental settings to train and evaluate the models on the two subsets of AVSBench separately.

\begin{table}[]
    \centering
    \setlength{\tabcolsep}{8pt}
    \begin{tabular}{l|cccc}
    \toprule
    \multicolumn{1}{c}{Subsets}  &  \multicolumn{2}{c}{S4} & \multicolumn{2}{c}{MS3} \\
     \cmidrule(r){2-3} \cmidrule(r){4-5} \multicolumn{1}{c }{Methods}   &   $\mathcal{M}_{\mathcal{J}}$ &  $\mathcal{M}_{\mathcal{F}}$ & $\mathcal{M}_{\mathcal{J}}$ & $\mathcal{M}_{\mathcal{F}}$ \\
    \midrule
SAM~\cite{kirillov2023segment}  & 55.08 &  0.739 & 53.96  & 0.638 \\
AV-SAM~\cite{mo2023av}  & 40.47 & 0.566  &  - & - \\
AP-SAM  & 69.61 & 0.796 & 51.58  & 0.578  \\
SAMA-AVS & \textbf{81.53} & \textbf{0.886}  &\textbf{63.14} & \textbf{0.691} \\
    \bottomrule
    \end{tabular}
    \vspace{-3pt}
    \caption{Comparison between different SAM-based methods on the testsets of S4 and MS3 subsets of AVSBench dataset.}
    \label{tab:comp_with_sams}
\vspace{-0.3cm}
\end{table}

The results are shown in Table~\ref{tab:comp_with_sams}.
On both subsets, SAMA-AVS outperforms other SAM-based methods by a large margin across all metrics.
On the S4 subset, we observe the AP-SAM achieves better performance than the vanilla SAM, and lags far behind SAMA-AVS. 
This may be attributed to the fact that, 
audio-visual fusion of AP-SAM at the decoding stage is insufficient, 
{\em i.e.}, decoder has only two cross-attention layers for fusion;
while SAMA-AVS enables more effective audio-visual fusion at the encoding stage with simple adapters.

\subsubsection{Comparison with Other SOTA methods.}\label{sec:com-sotas}
We also compare our proposed SAMA-AVS with the state-of-the-art AVS method~\cite{zhou2022avs}, which is based on an encoder-fusion-decoder framework. Additionally, we compare with other related audio-visual methods, 
including sound source localization (SSL) methods LVS~\cite{chen2021localizing} 
and MSSL~\cite{qian2020multiple}, video object segmentation (VOS) methods 3DC~\cite{mahadevan2020making} and SST~\cite{duke2021sstvos}, and salient object detection (SOD) methods iGAN~\cite{mao2021transformer} and LGVT~\cite{zhang2021learning}.

\begin{table}[]
    \small
    \centering
    \begin{tabular}{ll|cccc}
    \toprule
    \multicolumn{2}{c}{Subsets}  &  \multicolumn{2}{c}{S4} & \multicolumn{2}{c}{MS3} \\
     \cmidrule(r){3-4} \cmidrule(r){5-6} \multicolumn{2}{c }{Methods}   &   $\mathcal{M}_{\mathcal{J}}$ &  $\mathcal{M}_{\mathcal{F}}$ & $\mathcal{M}_{\mathcal{J}}$ & $\mathcal{M}_{\mathcal{F}}$ \\
    \midrule
    \multirow{2}{*}{SSL} & LVS~\cite{chen2021localizing} &37.9 &.510 &29.5 &  .330   \\ 
     & MSSL~\cite{qian2020multiple} & 44.9  & .663 &26.1 &.363 \\  \cline{2-6}
     \multirow{2}{*}{VOS} & 3DC~\cite{mahadevan2020making} & 57.1 & .759  & 36.9 & .503    \\ 
     & SST~\cite{duke2021sstvos} & 66.3 & .801 & 42.6 & .572 \\   \cline{2-6}
     \multirow{2}{*}{SOD} & iGAN~\cite{mao2021transformer} &61.6 & .778 & 42.9& .544    \\ 
     & LGVT~\cite{zhang2021learning} & 74.9 & .873& 40.7 & .593 \\  \cline{2-6}
     \multirow{2}{*}{AVS~\cite{zhou2022avs}} 
     & w/o F.T. & 78.7 &.879 & 54.0 & .645 \\ 
     & w/ F.T. & - & - & 51.45 & .671 \\ 
     \cline{2-6}
    \multirow{3}{*}{Ours}  & w/o F.T. & 81.53 & .886  & 63.14 & .691 \\
     & w/ F.T. & -  & -  & 66.30 & .730 \\
    & w/ F.T. (Syn.) & \textbf{83.17}  & \textbf{.901}  & \textbf{66.95} & \textbf{.754} \\
    \bottomrule
    \end{tabular}
    \vspace{-5pt}
    \caption{Comparison with the state-of-the-art method and other related audio-visual methods on test sets of two subsets of AVSBench dataset. (``F.T." denotes finetuning on the pretrained weights on S4; ``F.T.(Syn.)" denotes finetuning on the pretrained weights of model trained with AVS-Synthetic).}
    \label{tab:compare_with_sota}
\vspace{-0.3cm}
\end{table}

\begin{figure*}[h]
\centering
\vspace{0.1cm}
\includegraphics[width=.97\linewidth]{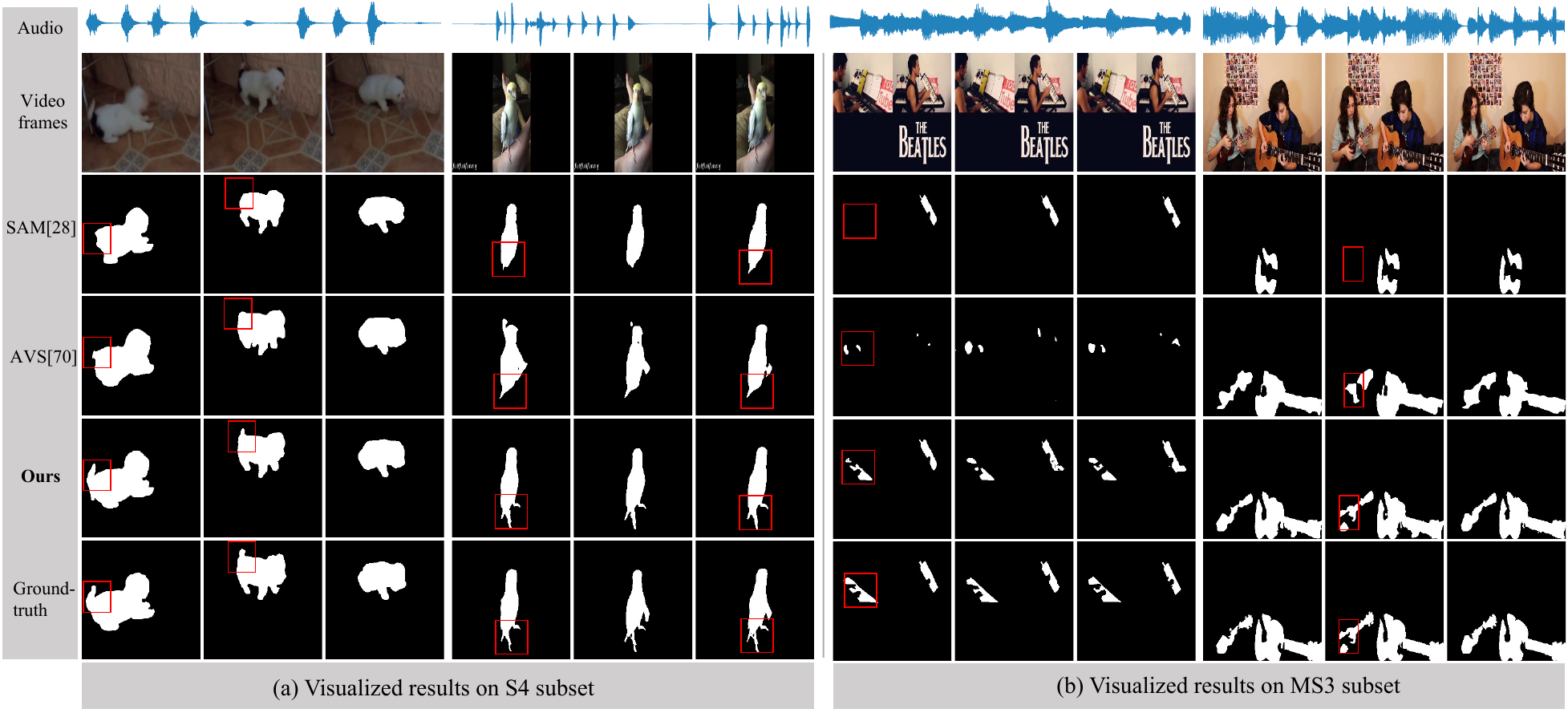}
\vspace{-0.1cm}
\caption{Qualitative results produced by our model and other methods on the test sets of S4 and MS3 subsets of AVSBench.}
\label{fig:qual-s4-ms3}
\vspace{-0.3cm}
\end{figure*}

According to the results presented in Table~\ref{tab:compare_with_sota}, 
our proposed approach outperforms existing methods in both subsets across all metrics, for example, achieving 81.53 for S4 subset and 63.14 for MS3 subset on $\mathcal{M}_{\mathcal{J}}$.
Note that, the results indicate that the performance of all methods is notably inferior on the MS3 subset as compared to the S4 subset. 
We conjecture that the causes stem from two aspects:
(i) the inherent difficulty of multi-sound segmentation;
(ii) the limited dataset size of MS3 subset.
To address this issue, we fine-tune the models that have already been trained on the S4 subset for MS3 subset.
And our method gives 3.16 \textit{improvement} on $\mathcal{M}_{\mathcal{J}}$ while AVS~\cite{zhou2022avs} has 2.55 \textit{drop}.
This shows that our method can effectively leverages the knowledge from the single sound object segmentation to improve the performance for multi-sound objects.

Lastly, we train the models for both S4 and MS3 of AVSBench dataset, by finetuning the weights pretrained from AVS-Synthetic dataset.
The models (w/F.T.(Syn.)) obtain further improvements which achieve 83.17 and 66.95 on $\mathcal{M}_{\mathcal{J}}$ for two subsets separately, outperforming current methods by a large margin. 
This shows the effectiveness of the proposed AVS-Synthetic dataset, implying the importance of data and model in tackling the AVS task.

\subsubsection{Ablation Study}\label{sec:abla}
Here we perform an ablation study to investigate the training objectives and the choice for the number of adapters in our proposed SAMA-AVS method.

\vspace{4pt}
\noindent \textbf{Training objectives.}
We train the models with either the BCE loss $\mathcal{L}_{BCE}$ or the IoU loss $\mathcal{L}_{IoU}$ on AVSBench S4 subset, and compare the results with the model trained with the combined loss.
The results in Table~\ref{tab:ablation-loss} shows that model trained with only BCE loss can already obtain satisfactory performance; by contrast, using the IoU loss only lead to very poor performance.
By combining the two objectives together, 
the model achieves the best performance.

\begin{table}[!htb]
\centering
\setlength{\tabcolsep}{8pt}
\begin{tabular}{cccc}
\toprule
Model with & $\mathcal{L}_{BCE}$ & $\mathcal{L}_{IoU}$ & $\mathcal{L}_{BCE}$ + $\mathcal{L}_{IoU}$   \\
\midrule
$\mathcal{M}_{\mathcal{J}}$ &  80.52   &  18.96    &  \textbf{81.53} \\
$\mathcal{M}_{\mathcal{F}}$ &  0.881   &  0.226   &  \textbf{0.886} \\
\bottomrule
\end{tabular}
\vspace{-4pt}
\caption{Comparison between the models trained with different training losses.}
\label{tab:ablation-loss}
\vspace{-0.4cm}
\end{table}

\vspace{3pt}
\noindent \textbf{Number of adapters.} 
We perform the experiments by varying the number of adapters in the model design using AVSBench S4 subset.
In the default SAMA-AVS design, the image encoder has 32 transformer encoder layers thus we insert an adapter before feeding the inputs for each of the layer, ending up with a total of 32 adapters.
We set the number of adapters to $\mathcal{N}\in \{2, 4, 8, 16, 32\} $, and evenly distribute $\mathcal{N}$ adapters among 32 transformer encoder layers. 
The results in Table~\ref{tab:ablation-num-ada} show that the performance of the model consistently improves as the number of adapters increases, demonstrating the importance of audio-visual fusion that the adapters facilitate.
\begin{table}[!htb]
\small
\centering
\begin{tabular}{cccccc}
\toprule
Adapter Num.  & 2 & 4 & 8 & 16 & 32  \\
\midrule
$\mathcal{M}_{\mathcal{J}}$ & 77.84    & 78.55  & 80.31    &  81.29 &  \textbf{81.53}   \\
$\mathcal{M}_{\mathcal{F}}$ &  .859   & .857  & .871    & .880  &  \textbf{.886}   \\
\bottomrule
\end{tabular}
\vspace{-4pt}
\caption{Comparison between the models trained with different numbers of adapters.}
\label{tab:ablation-num-ada}
\vspace{-0.5cm}
\end{table}

\subsubsection{Qualitative Examples}\label{sec:quali}
Figure~\ref{fig:qual-s4-ms3} illustrates some qualitative results produced by different methods on AVSBench.
The results show that on both subsets, 
the segmentations produced by our proposed SAMA-AVS best matches the ground-truths. Our model can precisely segment the whole body of the sounding objects without missing their parts; and it excels in handling the edges and details of the sounding target compared with other competitors, such as the tail of the dog, and the tail and feet of the bird, marked with red boxes in Figure~\ref{fig:qual-s4-ms3} (a).
For the MS3 setting, our model can also localise the multiple targets from the sound and segment them accurately.

\section{Conclusion}\label{sec:conclu}
In summary, this paper addresses the audio-visual segmentation task from two perspectives: training data and computational model.
For the former, we present a pipeline to synthesize audio-visual segmentation dataset by leveraging existing image segmentation datasets and audio dataset.
The proposed pipeline is \textit{annotation-free}, thus can be easily \textit{scaled up}. And as an instantiation of the pipeline, we collect the AVS-Synthetic dataset to facilitate the study of AVS problem. For the latter, we propose an architecture SAMA-AVS that aims to harness the ability of SAM foundation model with minimal training. 
Our proposed model introduces very few \textit{trainable} parameters (\textless 0.4\%) with the design of adapters to perform audio-visual fusion in the encoding stage, which is highly \textit{efficient} and \textit{effective}.
Extensive experimental results demonstrate the effectiveness of our proposed method, outperforming existing SAM-based methods and other competitors by a large margin.
By pretraining the proposed model on the synthetic dataset, the model further attains higher performance on both subsets of real AVSBench dataset. We hope our proposed dataset and method will inspire more research on the audio-visual understanding.

\noindent \textbf{Acknowledgements:} This work was supported by Shanghai Science and Technology Committee under Grant (No. 21511101100).

{\small
\bibliographystyle{ieee_fullname}
\bibliography{refs}

\begin{thebibliography}{10}\itemsep=-1pt

\bibitem{alayrac2022flamingo}
Jean-Baptiste Alayrac, Jeff Donahue, Pauline Luc, Antoine Miech, Iain Barr, Yana Hasson, Karel Lenc, Arthur Mensch, Katherine Millican, Malcolm Reynolds, et~al.
\newblock Flamingo: a visual language model for few-shot learning.
\newblock {\em Advances in Neural Information Processing Systems}, 35:23716--23736, 2022.

\bibitem{DBLP:journals/corr/abs-2304-12308}
Jiazhong Cen, Zanwei Zhou, Jiemin Fang, Wei Shen, Lingxi Xie, Dongsheng Jiang, Xiaopeng Zhang, and Qi Tian.
\newblock Segment anything in 3d with nerfs.
\newblock {\em CoRR}, abs/2304.12308, 2023.

\bibitem{chen2021localizing}
Honglie Chen, Weidi Xie, Triantafyllos Afouras, Arsha Nagrani, Andrea Vedaldi, and Andrew Zisserman.
\newblock Localizing visual sounds the hard way.
\newblock In {\em Proceedings of the IEEE/CVF Conference on Computer Vision and Pattern Recognition}, pages 16867--16876, 2021.

\bibitem{chen2020vggsound}
Honglie Chen, Weidi Xie, Andrea Vedaldi, and Andrew Zisserman.
\newblock Vggsound: A large-scale audio-visual dataset.
\newblock In {\em ICASSP 2020-2020 IEEE International Conference on Acoustics, Speech and Signal Processing (ICASSP)}, pages 721--725. IEEE, 2020.

\bibitem{DBLP:journals/corr/abs-2305-05803}
Tianle Chen, Zheda Mai, Ruiwen Li, and Wei{-}Lun Chao.
\newblock Segment anything model {(SAM)} enhanced pseudo labels for weakly supervised semantic segmentation.
\newblock {\em CoRR}, abs/2305.05803, 2023.

\bibitem{chen2023sam}
Tianrun Chen, Lanyun Zhu, Chaotao Ding, Runlong Cao, Shangzhan Zhang, Yan Wang, Zejian Li, Lingyun Sun, Papa Mao, and Ying Zang.
\newblock Sam fails to segment anything?--sam-adapter: Adapting sam in underperformed scenes: Camouflage, shadow, and more.
\newblock {\em arXiv preprint arXiv:2304.09148}, 2023.

\bibitem{chen2022vision}
Zhe Chen, Yuchen Duan, Wenhai Wang, Junjun He, Tong Lu, Jifeng Dai, and Yu Qiao.
\newblock Vision transformer adapter for dense predictions.
\newblock {\em arXiv preprint arXiv:2205.08534}, 2022.

\bibitem{DBLP:journals/corr/abs-2305-00035}
Dongjie Cheng, Ziyuan Qin, Zekun Jiang, Shaoting Zhang, Qicheng Lao, and Kang Li.
\newblock {SAM} on medical images: {A} comprehensive study on three prompt modes.
\newblock {\em CoRR}, abs/2305.00035, 2023.

\bibitem{duke2021sstvos}
Brendan Duke, Abdalla Ahmed, Christian Wolf, Parham Aarabi, and Graham~W Taylor.
\newblock Sstvos: Sparse spatiotemporal transformers for video object segmentation.
\newblock In {\em Proceedings of the IEEE/CVF Conference on Computer Vision and Pattern Recognition}, pages 5912--5921, 2021.

\bibitem{ephrat2018looking}
Ariel Ephrat, Inbar Mosseri, Oran Lang, Tali Dekel, Kevin Wilson, Avinatan Hassidim, William~T. Freeman, and Michael Rubinstein.
\newblock Looking to listen at the cocktail party: a speaker-independent audio-visual model for speech separation.
\newblock {\em {ACM} Trans. Graph.}, 37(4):112, 2018.

\bibitem{gao2021visualvoice}
Ruohan Gao and Kristen Grauman.
\newblock Visualvoice: Audio-visual speech separation with cross-modal consistency.
\newblock In {\em 2021 IEEE/CVF Conference on Computer Vision and Pattern Recognition (CVPR)}, pages 15490--15500. IEEE, 2021.

\bibitem{DBLP:journals/corr/abs-2306-00499}
Yifan Gao, Wei Xia, Dingdu Hu, and Xin Gao.
\newblock Desam: Decoupling segment anything model for generalizable medical image segmentation.
\newblock {\em CoRR}, abs/2306.00499, 2023.

\bibitem{gemmeke2017audio}
Jort~F Gemmeke, Daniel~PW Ellis, Dylan Freedman, Aren Jansen, Wade Lawrence, R~Channing Moore, Manoj Plakal, and Marvin Ritter.
\newblock Audio set: An ontology and human-labeled dataset for audio events.
\newblock In {\em 2017 IEEE international conference on acoustics, speech and signal processing (ICASSP)}, pages 776--780. IEEE, 2017.

\bibitem{gupta2019lvis}
Agrim Gupta, Piotr Dollar, and Ross Girshick.
\newblock Lvis: A dataset for large vocabulary instance segmentation.
\newblock In {\em Proceedings of the IEEE/CVF conference on computer vision and pattern recognition}, pages 5356--5364, 2019.

\bibitem{DBLP:journals/corr/abs-2305-11003}
Chunming He, Kai Li, Yachao Zhang, Guoxia Xu, Longxiang Tang, Yulun Zhang, Zhenhua Guo, and Xiu Li.
\newblock Weakly-supervised concealed object segmentation with sam-based pseudo labeling and multi-scale feature grouping.
\newblock {\em CoRR}, abs/2305.11003, 2023.

\bibitem{he2022masked}
Kaiming He, Xinlei Chen, Saining Xie, Yanghao Li, Piotr Doll{\'a}r, and Ross Girshick.
\newblock Masked autoencoders are scalable vision learners.
\newblock In {\em Proceedings of the IEEE/CVF Conference on Computer Vision and Pattern Recognition}, pages 16000--16009, 2022.

\bibitem{hershey2017cnn}
Shawn Hershey, Sourish Chaudhuri, Daniel~PW Ellis, Jort~F Gemmeke, Aren Jansen, R~Channing Moore, Manoj Plakal, Devin Platt, Rif~A Saurous, Bryan Seybold, et~al.
\newblock Cnn architectures for large-scale audio classification.
\newblock In {\em 2017 ieee international conference on acoustics, speech and signal processing (icassp)}, pages 131--135. IEEE, 2017.

\bibitem{houlsby2019parameter}
Neil Houlsby, Andrei Giurgiu, Stanislaw Jastrzebski, Bruna Morrone, Quentin De~Laroussilhe, Andrea Gesmundo, Mona Attariyan, and Sylvain Gelly.
\newblock Parameter-efficient transfer learning for nlp.
\newblock In {\em International Conference on Machine Learning}, pages 2790--2799. PMLR, 2019.

\bibitem{hu2019deep}
Di Hu, Feiping Nie, and Xuelong Li.
\newblock Deep multimodal clustering for unsupervised audiovisual learning.
\newblock In {\em Proceedings of the IEEE/CVF Conference on Computer Vision and Pattern Recognition}, pages 9248--9257, 2019.

\bibitem{hu2020discriminative}
Di Hu, Rui Qian, Minyue Jiang, Xiao Tan, Shilei Wen, Errui Ding, Weiyao Lin, and Dejing Dou.
\newblock Discriminative sounding objects localization via self-supervised audiovisual matching.
\newblock {\em Advances in Neural Information Processing Systems}, 33:10077--10087, 2020.

\bibitem{hu2021class}
Di Hu, Yake Wei, Rui Qian, Weiyao Lin, Ruihua Song, and Ji-Rong Wen.
\newblock Class-aware sounding objects localization via audiovisual correspondence.
\newblock {\em IEEE Transactions on Pattern Analysis and Machine Intelligence}, 2021.

\bibitem{ji2023sam}
Ge{-}Peng Ji, Deng{-}Ping Fan, Peng Xu, Ming{-}Ming Cheng, Bowen Zhou, and Luc~Van Gool.
\newblock {SAM} struggles in concealed scenes - empirical study on "segment anything".
\newblock {\em CoRR}, abs/2304.06022, 2023.

\bibitem{DBLP:journals/corr/abs-2304-05750}
Wei Ji, Jingjing Li, Qi Bi, Wenbo Li, and Li Cheng.
\newblock Segment anything is not always perfect: An investigation of {SAM} on different real-world applications.
\newblock {\em CoRR}, abs/2304.05750, 2023.

\bibitem{DBLP:journals/corr/abs-2305-01275}
Peng{-}Tao Jiang and Yuqi Yang.
\newblock Segment anything is {A} good pseudo-label generator for weakly supervised semantic segmentation.
\newblock {\em CoRR}, abs/2305.01275, 2023.

\bibitem{karimi2021compacter}
Rabeeh Karimi~Mahabadi, James Henderson, and Sebastian Ruder.
\newblock Compacter: Efficient low-rank hypercomplex adapter layers.
\newblock {\em Advances in Neural Information Processing Systems}, 34:1022--1035, 2021.

\bibitem{kazakos2019epic}
Evangelos Kazakos, Arsha Nagrani, Andrew Zisserman, and Dima Damen.
\newblock Epic-fusion: Audio-visual temporal binding for egocentric action recognition.
\newblock In {\em Proceedings of the IEEE/CVF International Conference on Computer Vision}, pages 5492--5501, 2019.

\bibitem{kirillov2023segment}
Alexander Kirillov, Eric Mintun, Nikhila Ravi, Hanzi Mao, Chloe Rolland, Laura Gustafson, Tete Xiao, Spencer Whitehead, Alexander~C Berg, Wan-Yen Lo, et~al.
\newblock Segment anything.
\newblock {\em arXiv preprint arXiv:2304.02643}, 2023.

\bibitem{kuznetsova2020open}
Alina Kuznetsova, Hassan Rom, Neil Alldrin, Jasper Uijlings, Ivan Krasin, Jordi Pont-Tuset, Shahab Kamali, Stefan Popov, Matteo Malloci, Alexander Kolesnikov, et~al.
\newblock The open images dataset v4: Unified image classification, object detection, and visual relationship detection at scale.
\newblock {\em International Journal of Computer Vision}, 128(7):1956--1981, 2020.

\bibitem{lee2020cross}
Jun-Tae Lee, Mihir Jain, Hyoungwoo Park, and Sungrack Yun.
\newblock Cross-attentional audio-visual fusion for weakly-supervised action localization.
\newblock In {\em International Conference on Learning Representations}, 2020.

\bibitem{lee2020parameter}
Sangho Lee, Youngjae Yu, Gunhee Kim, Thomas Breuel, Jan Kautz, and Yale Song.
\newblock Parameter efficient multimodal transformers for video representation learning.
\newblock {\em arXiv preprint arXiv:2012.04124}, 2020.

\bibitem{li2022exploring}
Yanghao Li, Hanzi Mao, Ross Girshick, and Kaiming He.
\newblock Exploring plain vision transformer backbones for object detection.
\newblock In {\em Computer Vision--ECCV 2022: 17th European Conference, Tel Aviv, Israel, October 23--27, 2022, Proceedings, Part IX}, pages 280--296. Springer, 2022.

\bibitem{lin2021unsupervised}
Yan{-}Bo Lin, Hung{-}Yu Tseng, Hsin{-}Ying Lee, Yen{-}Yu Lin, and Ming{-}Hsuan Yang.
\newblock Unsupervised sound localization via iterative contrastive learning.
\newblock {\em CoRR}, abs/2104.00315, 2021.

\bibitem{lin2019dual}
Yan-Bo Lin, Yu-Jhe Li, and Yu-Chiang~Frank Wang.
\newblock Dual-modality seq2seq network for audio-visual event localization.
\newblock In {\em ICASSP 2019-2019 IEEE International Conference on Acoustics, Speech and Signal Processing (ICASSP)}, pages 2002--2006. IEEE, 2019.

\bibitem{lin2023vision}
Yan-Bo Lin, Yi-Lin Sung, Jie Lei, Mohit Bansal, and Gedas Bertasius.
\newblock Vision transformers are parameter-efficient audio-visual learners.
\newblock In {\em Proceedings of the IEEE/CVF Conference on Computer Vision and Pattern Recognition}, pages 2299--2309, 2023.

\bibitem{lin2022frozen}
Ziyi Lin, Shijie Geng, Renrui Zhang, Peng Gao, Gerard de Melo, Xiaogang Wang, Jifeng Dai, Yu Qiao, and Hongsheng Li.
\newblock Frozen clip models are efficient video learners.
\newblock In {\em Computer Vision--ECCV 2022: 17th European Conference, Tel Aviv, Israel, October 23--27, 2022, Proceedings, Part XXXV}, pages 388--404. Springer, 2022.

\bibitem{liu2022few}
Haokun Liu, Derek Tam, Mohammed Muqeeth, Jay Mohta, Tenghao Huang, Mohit Bansal, and Colin~A Raffel.
\newblock Few-shot parameter-efficient fine-tuning is better and cheaper than in-context learning.
\newblock {\em Advances in Neural Information Processing Systems}, 35:1950--1965, 2022.

\bibitem{liu2022exploiting}
Jinxiang Liu, Chen Ju, Weidi Xie, and Ya Zhang.
\newblock Exploiting transformation invariance and equivariance for self-supervised sound localisation.
\newblock In {\em Proceedings of the 30th ACM International Conference on Multimedia}, pages 3742--3753, 2022.

\bibitem{liu2023explicit}
Weihuang Liu, Xi Shen, Chi-Man Pun, and Xiaodong Cun.
\newblock Explicit visual prompting for low-level structure segmentations.
\newblock In {\em Proceedings of the IEEE/CVF Conference on Computer Vision and Pattern Recognition}, pages 19434--19445, 2023.

\bibitem{DBLP:journals/corr/abs-2306-09347}
Youquan Liu, Lingdong Kong, Jun Cen, Runnan Chen, Wenwei Zhang, Liang Pan, Kai Chen, and Ziwei Liu.
\newblock Segment any point cloud sequences by distilling vision foundation models.
\newblock {\em CoRR}, abs/2306.09347, 2023.

\bibitem{DBLP:journals/corr/abs-2305-13310}
Yang Liu, Muzhi Zhu, Hengtao Li, Hao Chen, Xinlong Wang, and Chunhua Shen.
\newblock Matcher: Segment anything with one shot using all-purpose feature matching.
\newblock {\em CoRR}, abs/2305.13310, 2023.

\bibitem{loshchilov2017decoupled}
Ilya Loshchilov and Frank Hutter.
\newblock Decoupled weight decay regularization.
\newblock {\em arXiv preprint arXiv:1711.05101}, 2017.

\bibitem{DBLP:journals/corr/abs-2304-12306}
Jun Ma and Bo Wang.
\newblock Segment anything in medical images.
\newblock {\em CoRR}, abs/2304.12306, 2023.

\bibitem{mahadevan2020making}
Sabarinath Mahadevan, Ali Athar, Aljosa Osep, Laura Leal{-}Taix{\'{e}}, Bastian Leibe, and Sebastian Hennen.
\newblock Making a case for 3d convolutions for object segmentation in videos.
\newblock In {\em 31st British Machine Vision Conference 2020, {BMVC} 2020, Virtual Event, UK, September 7-10, 2020}. {BMVA} Press, 2020.

\bibitem{mao2021transformer}
Yuxin Mao, Jing Zhang, Zhexiong Wan, Yuchao Dai, Aixuan Li, Yunqiu Lv, Xinyu Tian, Deng{-}Ping Fan, and Nick Barnes.
\newblock Transformer transforms salient object detection and camouflaged object detection.
\newblock {\em CoRR}, abs/2104.10127, 2021.

\bibitem{mo2023av}
Shentong Mo and Yapeng Tian.
\newblock Av-sam: Segment anything model meets audio-visual localization and segmentation.
\newblock {\em arXiv preprint arXiv:2305.01836}, 2023.

\bibitem{owens2018audio}
Andrew Owens and Alexei~A Efros.
\newblock Audio-visual scene analysis with self-supervised multisensory features.
\newblock In {\em Proceedings of the European Conference on Computer Vision (ECCV)}, pages 631--648, 2018.

\bibitem{qian2020multiple}
Rui Qian, Di Hu, Heinrich Dinkel, Mengyue Wu, Ning Xu, and Weiyao Lin.
\newblock Multiple sound sources localization from coarse to fine.
\newblock In {\em European Conference on Computer Vision}, pages 292--308. Springer, 2020.

\bibitem{radford2021learning}
Alec Radford, Jong~Wook Kim, Chris Hallacy, Aditya Ramesh, Gabriel Goh, Sandhini Agarwal, Girish Sastry, Amanda Askell, Pamela Mishkin, Jack Clark, et~al.
\newblock Learning transferable visual models from natural language supervision.
\newblock In {\em International conference on machine learning}, pages 8748--8763. PMLR, 2021.

\bibitem{rebuffi2018efficient}
Sylvestre-Alvise Rebuffi, Hakan Bilen, and Andrea Vedaldi.
\newblock Efficient parametrization of multi-domain deep neural networks.
\newblock In {\em Proceedings of the IEEE Conference on Computer Vision and Pattern Recognition}, pages 8119--8127, 2018.

\bibitem{rouditchenko2019self}
Andrew Rouditchenko, Hang Zhao, Chuang Gan, Josh McDermott, and Antonio Torralba.
\newblock Self-supervised audio-visual co-segmentation.
\newblock In {\em ICASSP 2019-2019 IEEE International Conference on Acoustics, Speech and Signal Processing (ICASSP)}, pages 2357--2361. IEEE, 2019.

\bibitem{senocak2018learning}
Arda Senocak, Tae-Hyun Oh, Junsik Kim, Ming-Hsuan Yang, and In~So Kweon.
\newblock Learning to localize sound source in visual scenes.
\newblock In {\em Proceedings of the IEEE Conference on Computer Vision and Pattern Recognition}, pages 4358--4366, 2018.

\bibitem{DBLP:journals/corr/abs-2306-06370}
Tal Shaharabany, Aviad Dahan, Raja Giryes, and Lior Wolf.
\newblock Autosam: Adapting {SAM} to medical images by overloading the prompt encoder.
\newblock {\em CoRR}, abs/2306.06370, 2023.

\bibitem{sharir2021image}
Gilad Sharir, Asaf Noy, and Lihi Zelnik-Manor.
\newblock An image is worth 16x16 words, what is a video worth?
\newblock {\em arXiv preprint arXiv:2103.13915}, 2021.

\bibitem{song2022self}
Zengjie Song, Yuxi Wang, Junsong Fan, Tieniu Tan, and Zhaoxiang Zhang.
\newblock Self-supervised predictive learning: A negative-free method for sound source localization in visual scenes.
\newblock In {\em Proceedings of the IEEE/CVF Conference on Computer Vision and Pattern Recognition}, pages 3222--3231, 2022.

\bibitem{stickland2019bert}
Asa~Cooper Stickland and Iain Murray.
\newblock Bert and pals: Projected attention layers for efficient adaptation in multi-task learning.
\newblock In {\em International Conference on Machine Learning}, pages 5986--5995. PMLR, 2019.

\bibitem{sung2022lst}
Yi-Lin Sung, Jaemin Cho, and Mohit Bansal.
\newblock Lst: Ladder side-tuning for parameter and memory efficient transfer learning.
\newblock {\em arXiv preprint arXiv:2206.06522}, 2022.

\bibitem{sung2022vl}
Yi-Lin Sung, Jaemin Cho, and Mohit Bansal.
\newblock Vl-adapter: Parameter-efficient transfer learning for vision-and-language tasks.
\newblock In {\em Proceedings of the IEEE/CVF Conference on Computer Vision and Pattern Recognition}, pages 5227--5237, 2022.

\bibitem{tang2023can}
Lv Tang, Haoke Xiao, and Bo Li.
\newblock Can sam segment anything? when sam meets camouflaged object detection.
\newblock {\em arXiv preprint arXiv:2304.04709}, 2023.

\bibitem{tian2020unified}
Yapeng Tian, Dingzeyu Li, and Chenliang Xu.
\newblock Unified multisensory perception: Weakly-supervised audio-visual video parsing.
\newblock In {\em European Conference on Computer Vision}, pages 436--454. Springer, 2020.

\bibitem{tzinis2022audioscopev2}
Efthymios Tzinis, Scott Wisdom, Tal Remez, and John~R. Hershey.
\newblock Audioscopev2: Audio-visual attention architectures for calibrated open-domain on-screen sound separation.
\newblock In Shai Avidan, Gabriel~J. Brostow, Moustapha Ciss{\'{e}}, Giovanni~Maria Farinella, and Tal Hassner, editors, {\em Computer Vision - {ECCV} 2022 - 17th European Conference, Tel Aviv, Israel, October 23-27, 2022, Proceedings, Part {XXXVII}}, volume 13697 of {\em Lecture Notes in Computer Science}, pages 368--385. Springer, 2022.

\bibitem{DBLP:journals/corr/abs-2304-14674}
An Wang, Mobarakol Islam, Mengya Xu, Yang Zhang, and Hongliang Ren.
\newblock {SAM} meets robotic surgery: An empirical study in robustness perspective.
\newblock {\em CoRR}, abs/2304.14674, 2023.

\bibitem{DBLP:journals/corr/abs-2305-16698}
Yonghui Wang, Wengang Zhou, Yunyao Mao, and Houqiang Li.
\newblock Detect any shadow: Segment anything for video shadow detection.
\newblock {\em CoRR}, abs/2305.16698, 2023.

\bibitem{DBLP:journals/corr/abs-2304-12620}
Junde Wu, Rao Fu, Huihui Fang, Yuanpei Liu, Zhaowei Wang, Yanwu Xu, Yueming Jin, and Tal Arbel.
\newblock Medical {SAM} adapter: Adapting segment anything model for medical image segmentation.
\newblock {\em CoRR}, abs/2304.12620, 2023.

\bibitem{zhang2021learning}
Jing Zhang, Jianwen Xie, Nick Barnes, and Ping Li.
\newblock Learning generative vision transformer with energy-based latent space for saliency prediction.
\newblock {\em Advances in Neural Information Processing Systems}, 34:15448--15463, 2021.

\bibitem{DBLP:journals/corr/abs-2305-03048}
Renrui Zhang, Zhengkai Jiang, Ziyu Guo, Shilin Yan, Junting Pan, Hao Dong, Peng Gao, and Hongsheng Li.
\newblock Personalize segment anything model with one shot.
\newblock {\em CoRR}, abs/2305.03048, 2023.

\bibitem{DBLP:journals/corr/abs-2306-06113}
Xiaofeng Zhang, Chaochen Gu, and Shanying Zhu.
\newblock Sam-helps-shadow: When segment anything model meet shadow removal.
\newblock {\em CoRR}, abs/2306.06113, 2023.

\bibitem{zhao2019sound}
Hang Zhao, Chuang Gan, Wei-Chiu Ma, and Antonio Torralba.
\newblock The sound of motions.
\newblock In {\em Proceedings of the IEEE/CVF International Conference on Computer Vision}, pages 1735--1744, 2019.

\bibitem{zhao2018sound}
Hang Zhao, Chuang Gan, Andrew Rouditchenko, Carl Vondrick, Josh McDermott, and Antonio Torralba.
\newblock The sound of pixels.
\newblock In {\em Proceedings of the European conference on computer vision (ECCV)}, pages 570--586, 2018.

\bibitem{zhou2016learning}
Bolei Zhou, Aditya Khosla, Agata Lapedriza, Aude Oliva, and Antonio Torralba.
\newblock Learning deep features for discriminative localization.
\newblock In {\em Proceedings of the IEEE conference on computer vision and pattern recognition}, pages 2921--2929, 2016.

\bibitem{zhou2022avs}
Jinxing Zhou, Jianyuan Wang, Jiayi Zhang, Weixuan Sun, Jing Zhang, Stan Birchfield, Dan Guo, Lingpeng Kong, Meng Wang, and Yiran Zhong.
\newblock Audio-visual segmentation.
\newblock In {\em European Conference on Computer Vision}, 2022.

\end{thebibliography}
}

\end{document}